\documentclass{article}

\usepackage{microtype}
\usepackage{graphicx}
\usepackage{subfigure}
\usepackage{multirow}
\usepackage{booktabs} 

\usepackage{hyperref}

\usepackage[accepted]{icml2024}

\usepackage{amsmath}
\usepackage{amssymb}
\usepackage{mathtools}
\usepackage{amsthm}

\usepackage[capitalize,noabbrev]{cleveref}

\usepackage[normalem]{ulem}

\usepackage[textwidth=1.2in,textsize=tiny]{todonotes}

\newcommand{\ba}[1]{\begin{align}#1\end{align}}

\newcommand{\minus}{\scalebox{0.5}[1.0]{$-$}}

\usepackage{stackengine}

\makeatletter
\newcommand{\distas}[1]{\mathbin{\overset{#1}{\kern\z@\sim}}}%

\newcommand{\beqs}{\vspace{0mm}\begin{eqnarray}}
\newcommand{\eeqs}{\vspace{0mm}\end{eqnarray}}
\newcommand{\barr}{\begin{array}}
\newcommand{\earr}{\end{array}}

\newcommand{\sv}[0]{{\boldsymbol{s}}}
\newcommand{\tv}[0]{{\boldsymbol{t}} }

\newcommand{\xv}{\boldsymbol{x}}

\newcommand{\ours}{{SD}}
\usepackage{amssymb}
\usepackage{pifont}
\newcommand{\cmark}{\ding{51}}%
\usepackage[textsize=tiny]{todonotes}

\icmltitlerunning{Switchable Decision: Dynamic Neural Generation Networks}

\begin{document}

\twocolumn[
\icmltitle{Switchable Decision: Dynamic Neural Generation Networks
}




\begin{icmlauthorlist}
\icmlauthor{Shujian Zhang}{to}
\icmlauthor{Korawat Tanwisuth}{to}
\icmlauthor{Chengyue Gong}{to}
\icmlauthor{Pengcheng He}{}
\icmlauthor{Mingyuan Zhou}{to}
\end{icmlauthorlist}

\icmlaffiliation{to}{The University of Texas at Austin}

\icmlcorrespondingauthor{Shujian Zhang}{szhang19@utexas.edu}
\icmlkeywords{Machine Learning, ICML}

\vskip 0.3in
]



\printAffiliationsAndNotice{}  

\begin{abstract} 
Auto-regressive generation models achieve competitive performance across many different NLP tasks such as summarization, question answering, and classifications. However, they
are also known for being slow in inference,
which makes them challenging to deploy in real-time applications. We propose a switchable decision to accelerate inference by dynamically assigning computation resources for each data instance. 
Automatically making decisions on where to skip and how to balance quality and computation cost with constrained optimization, our dynamic neural generation networks enforce the efficient inference path and determine the optimized trade-off. 
Experiments across question answering, summarization, and classification benchmarks show that our method benefits from less computation cost during inference while keeping the same accuracy.
Extensive experiments and ablation studies demonstrate that our method can be general, effective, and beneficial for many NLP tasks.

\end{abstract}

\section{Introduction}
Large-scale pre-trained language models such as BART \cite{lewis2019bart} have demonstrated a significant performance gain to the natural language processing (NLP) community but generally come with the cost of a heavy computational burden. 
Besides pre-training and fine-tuning, inference of such a large model also comes with a heavy computational cost.
On IoT (Internet of things) devices and real-world applications, lower computation cost tolerance and restricted computation resource during inference impede these models from deployment.

Recent efforts of efficient inference mainly focus on pruning or compressing the model parameters, \emph{e.g.}, pruning unimportant parts of the neural model weights \cite{han2015learning, fan2019reducing, gordon2020compressing}, quantizing the number
of bits needed \cite{lin2016fixed, shen2020q}, distilling from large teacher
models to small student models \cite{hinton2015distilling, jiao2019tinybert}. These methods 
produce only one small model with
a predetermined target size.
Another direction is to switch the model parameters for different data instances, \emph{e.g.}, the mixture of experts \cite{shazeer2017outrageously}, and switch transformer \cite{fedus2021switch}. 
Early exiting, which adaptively produces a series of small
models for different data instances, is one of the most common practices. 
Most previous work makes exit decisions based on either the confidence of output probability distributions or a trained agent. 
In this work, we propose a carefully designed candidate space for encoder-decoder auto-regressive models and enhance the optimization strategies when training the agent.

In this spirit, we explore the problem of dynamically allocating computation across a generation model. In particular, we consider a standard encoder-decoder transformer auto-regressive generation model. 
It comprises a stacked structure with multiple layers,
each having a multi-head attention layer
followed by a feed-forward network (FFN) layer \cite{zhang2021alignment, zhang2021bayesian, dai2022one, tanwisuth2023pouf}.
To this end, we introduce a dynamic neural network for the auto-regressive generation models,  which includes the attention, feed-forward, and input sequence as the candidate space for switchable decisions.
Our method generates an input-dependent inference strategy for each data. 
For each input sequence, the reinforcement learning agent outputs all the  decisions for skipping or keeping each candidate. 
With the first-layer hidden representations as the input, 
the policy network is trained to maximize a reward that incentives the use of as few blocks or tokens as possible while preserving the prediction accuracy.

We propose learning optimal switchable strategies that simultaneously preserve prediction accuracy and minimal computation usage based on input-specific decisions. The constrained optimization is utilized 
as a more principled approach for trading off these two targets (quality \emph{v.s.} efficiency).
We target keeping the predicted quality while achieving better efficiency as far as possible.
A gradient-based constrained optimization algorithm is implemented under our framework.

We run extensive experiments across summarization, \emph{e.g.}, XSum \cite{narayan2018don} and CNN/DM  \cite{hermann2015teaching}, 
question answering, \emph{e.g.}, SQuAD 1.1 \citep{rajpurkar2016squad} and SQuAD 2.0 \citep{rajpurkar2018know}), and GLUE \cite{wang2018glue} classification tasks. 
\ding{182} Our method not only shows comparable performance across different tasks and datasets but also accelerates model inference by up to 40\% with negligible model quality degradation. 
\ding{183} Furthermore, we provide extensive ablation studies on different design choices for the proposed method, including  the encoder-only or decoder-only switchable schemes. 
\ding{184} Our analysis shows the switchable decision contributes the efficiency improvement and accuracy consistency, helping the generation model to choose the inference path and candidates dynamically. 
\ding{185} To the best of our knowledge, we present the first switchable decision in the language generation model setting by dynamically making the inference decisions in summarization, question answering, and classification.
Our contributions are summarized as follows:
\begin{itemize}
\setlength{\itemsep}{0pt}
\setlength{\parsep}{0pt}
\setlength{\parskip}{0pt}
    \item Present a dynamic network for switchable decisions embracing
    attention, feed-forward, and input sequence as skipping candidates. 
    \item Propose an efficient and effective way to train the skipping strategies, which can optimize the trade-off between computation  and quality.
    \item Verify the effectiveness and general applicability of the proposed method in various NLP tasks, \emph{$e.g.$}, summarization, question answering, and classification benchmarks, and provide a rich analysis of our method with various design choices. 
\end{itemize}

\section{Related Work and Background}
\paragraph{Compact Network Design and Model Compression}
For model compression, pruning  removes unimportant parts of the neural network \cite{han2015deep, fan2019reducing, gordon2020compressing}, 
quantization   targets the number of bits needed to operate a neural network \cite{shen2020q}, and distillation  transfers knowledge from large teacher models to small student models \cite{chen2017learning, jiao2019tinybert}. Efficient network architectures such as MobileBERT \cite{sun2020mobilebert} and ALBERT \cite{Lan2020ALBERTAL} have also been
explored for lightweight neural network architectures.
Compared to these previous approaches, we focus on dynamic networks 
with the switchable candidate design to best reduce total computation without degrading prediction accuracy.

\paragraph{Dynamic Networks}
Dynamic networks enable adaptive computation for various input instances that have been conducted for natural language tasks.
Text skimming \cite{campos2017skip, hansen2019neural}
learns to skip state updates and shortens the effective
size of the computational graph. Dynamic jumping \cite{yu2018fast, fu2018speed} strategically
skips some tokens without reading them, and directly
jumps to an arbitrary location.
Early exiting for pretrained models has been explored by previous literature. 
RTJ \cite{schwartz2020right}, DeeBERT \cite{xin2020deebert}, and FastBERT \cite{liu2020fastbert} make early exiting decisions based on confidence (or its variants) of the predicted probability distribution and are therefore limited to classification tasks. PABEE \cite{zhou2020bert} and BERxiT \cite{xin2021berxit}
propose patience-based early exiting by exploiting the layer information.
Runtime Neural Pruning \cite{lin2017runtime}, SkipNet \cite{wang2018skipnet}, and BlockDrop \cite{wu2018blockdrop} use reinforcement learning (RL) to decide whether to execute a network module. 
Inspired by them, we incorporate lightweight reinforcement learning to make input-dependent decisions and  build a diversified switchable candidate space. With the constrained optimization approach, our method saves computational costs without loss of accuracy.

\begin{figure*}[t]
\centering
\includegraphics[width=15.0cm]{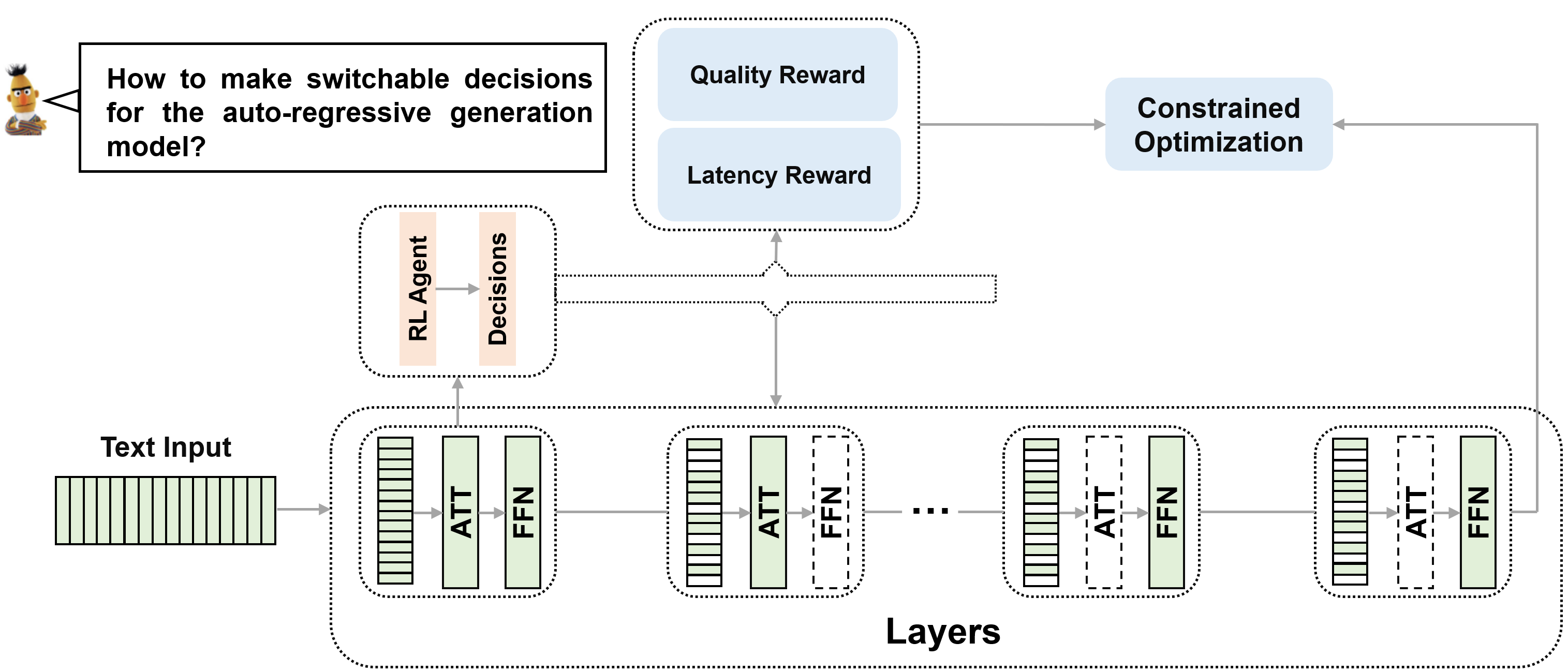}
\caption{Overview of the dynamic network. Some notations are labeled along with corresponding components. `Layers' refers to layers within the auto-regressive generation model.
`ATT' refers to the attention candidate, `FFN' refers to the feed-forward candidate, `Text Input' refers to the token candidate, and `Decisions' refers to the skipping decisions from the reinforcement learning agent.
The green color represents not skipping. The no-fill in the text input and the dashed line with the no-fill color box represents the skipping.
}
\label{fig:pipeline}
\end{figure*}

\section{Method} \label{sec:method_section}
Our switchable decision (Figure \ref{fig:pipeline}) network focuses on speeding up the inference time for an autoregressive language generation model. Specifically, we suggest a general recipe for the switchable decision: 1) construct the versatile decision space, 2) utilize the input-dependent reinforcement learning agent, and 3) propose the lexicographic (lexico) optimization strategy. 



Denote input $\mathbf{o}$ = $(\mathbf{o}_0, \cdots, \mathbf{o}_n)$.
With a series of $n$ tokens, a transformer-based language generation model, $\mathcal{M}$ with $L$ layers, first embeds the tokens to form a matrix $O_e \in \mathbf{R}^{n\times e}$, where $e$ is the dimension of the embedding space.
These token representations then go through the encoders and decoders of the language model. To speed up the inference time while maintaining similar high quality, we decide whether each input data should skip one layer. This decision problem grows exponentially as we increase the number of layers. Moreover, because of the discrete nature of the decision space, optimization becomes challenging. In this section, we outline our unique design choices to accomplish our goal and overcome optimization challenges.

\subsection{Construct Discrete Decision Space} \label{sec:construct_decision_space}
We propose learning the best configurations of \emph{(input, inference paths)} pair for each example using a switchable decision network to speed up inference time. We consider three search space candidates, namely, the attention layer, the feed-forward layer, and query inputs after the first layer. We now explain the details of each search space below.

\paragraph{Attention Candidate.}
A key component of a transformer-based language model is the attention layer. \citet{zhang2019all} discover that some layers are redundant. To decide whether to skip a certain layer, we model these decisions as a sequence of i.i.d. Bernoulli random variables parameterized by a policy network $q$. Let $\mathbf{b}_{l}$ denote the switchable decision of the $l^\mathrm{th}$ layer, defined as
\begin{align}
   \mathbf{b}_{l}=\begin{cases}
          1 \quad &\text{with probability}\quad \, g(\xv)_l   \\
          0 \quad &\text{with probability}\quad \, 1-g(\xv)_l \\
     \end{cases},
\end{align}
where $\xv \in \mathbf{R}^e$ denotes the input of the decision unit, 
and we apply the first encoder layer output as $\xv$.
The policy network, $g$, learns instance-specific probabilities of keeping the hidden representations of each layer. To perform skipping, we sample from this distribution and broadcast the indicators, $\mathbf{b}^\mathrm{att}_{l}$, to the input representations of attention layers. 

\paragraph{Feed-Forward Candidate.}
In the same spirit, the feed-forward layers may contain redundant information. Thus, we consider skipping these layers using the same approach as that done in the attention. We decide whether to skip or not based on the indicator $\mathbf{b}^\mathrm{ffn}_{l}$.
The design of the policy network is the same as that of the attention layer.

\paragraph{Token Candidate.}
In addition to skipping the layers, 
skipping the tokens can also be an alternative way to save computation costs. We create two token skipping strategies: \ding{192} skipping the last $p\%$ tokens and \ding{193} uniformly skipping $p\%$ tokens. For the former, we set $p$ to $10$, $20$, and $30$. For the latter, $p$ is equal to $25$, $33$, and $50$. To decide which strategy to use, we optimize a categorical random variable parameterized by a function $h(\cdot)$. 
The input of $h(\cdot)$ is the same as $g(\cdot)$, and the output of $h(\cdot)$ is a distribution over all six candidate decisions.

\paragraph{Encoder and Decoder Structure.}
Our interested architecture contains encoders and decoders.
For the encoders, we apply attention skipping and feed-forward skipping together with token skipping.
For the decoders, since every token is meaningful for the final outputs, we only apply attention skipping and feed-forward skipping.
When making decisions, we sample from the 
outputs of our policy network, and broadcast the decisions to the hidden representations of each layer. 

\subsection{Reinforcement Learning Agent}
\paragraph{Policy Network Architecture.}  
Since we aim to speed up the inference process, a simple design for the policy network is adopted. We utilize a one-layer MLP with layer normalization and ReLU activation function. To output a Binomial 
distribution over decisions, we apply the sigmoid activation to the outputs of the network for attention and feed-forward candidates. We use the softmax function to output the distribution over the choices for token candidates.

\paragraph{Parameterization.} 
During the training process, we sample from the decision distributions, which are parameterized by the policy network. The distribution of the switchable decisions for the layers can be represented as a $2L$-dimensional Bernoulli distribution, which can be written as:
\begin{align}
\label{eq:prob:layer}
\pi(\sv \mid \xv)= \prod_{l=1}^{2L} g_l(\xv)^{s_{l}}(1-g_l(\xv))^{1- s_{l}},
\end{align}
where $\sv = \{\mathbf{b}_{l}^\mathrm{att}\}_{l=1}^L \bigcup \{\mathbf{b}_{l}^\mathrm{ffn}\}_{l=1}^L$. Similarly, the distribution of the token skipping decisions can be represented as a categorical distribution, which can be formalized as:
\begin{align}
\label{eq:prob:token}
\eta(a \mid \xv)= \prod_{j=1}^J h_j(\xv)^{\mathbf{1} (a=j)},
\end{align}
where $a$ denotes the choice of the skipping strategy, and $J$ indicates the total number of strategies. We apply seven candidates in practice.

\paragraph{Reward.}
We define the reward function \cite{yang2022unified, yang2022regularizing, feng2023fantastic} as a trade-off between quality and computational cost. Given an inference path and a data instance, the reward can be computed from the computation (estimated FLOPs). Intuitively skipping layers will have high reward. We further refer quality as accuracy and loss in the following way:
\ba{
   R(\sv, a)= \mathrm{quality} + \lambda \mathrm{computation}, 
   \label{eq:our_reward}
}
where $\mathrm{quality}$ is $\mathrm{-loss}$, $\mathrm{computation}$ is the estimated FLOPs (floating point operations), and $\lambda$ is a coefficient.
The overall loss function is defined as the expected value of the reward:

\begin{align}
J=\mathbf{E}_{\sv \sim \pi, ~ \tv \sim \eta}[R(\sv, a)],
\end{align}
where $\pi$ and $\eta$ are defined in \eqref{eq:prob:layer} and \eqref{eq:prob:token}, respectively.

\paragraph{Optimization.}  
To optimize our policy network, we apply policy gradient to compute the gradient of $J$, and update the parameters of the policy network. We 
use a self-critical baseline to reduce the variance of the gradients. 
The constraint-optimization strategy is further applied on the quality and computation. Details are in the next section.

\paragraph{During Inference.}
Unlike the training process, we do not sample the skipping decisions during inference. Instead, we choose the decisions which maximize the likelihood function. 


\subsection{Constrained Optimization}
\paragraph{Trade-off is a Problem.} 
In the joint training of the main network and the policy network, a trade-off between
quality and computation is important. The linear combination of multiple objectives is the most widely used approach. However, the coefficient of the combination requires manual tuning, and it is  theoretically unsuitable for non-convex functions. In this work, we consider constrained optimization on trading off two objectives, with a special emphasis on lexicographic (lexico) optimization.

\paragraph{Our Equation.}
To optimize the trade-off between quality and computation in Eqn \eqref{eq:our_reward}, we propose to use lexicographic optimization, in which the parameters are iteratively updated as 
\ba{
\theta_{t+1} \leftarrow \theta_t-\gamma_t e_t,
\label{eq:our_pareto}
}
where $\gamma_t \geq  0$ is an adaptive step size and $e_t \in \mathbb{R}^{d}$ is an update direction to be chosen to balance the minimization of $f$ and constraint satisfaction on $q$.
One of the objectives (say $f$ which is $\mathrm{computation}$ in our case) is of secondary
importance w.r.t. the other one (say $q$  which is $\mathrm{quality}$). 
The design criterion for the constrained optimization is when the constraint is not satisfied (i.e., $q(\theta_t) \ge c$), the focus becomes decreasing $q$ to satisfy the constraint as soon as possible; in the meantime, $f$ performs as a secondary objective indicating that $f$ should be minimized to the degree that it does not hurt the descent of $q$.
Therefore, we apply the following update rule to obtain such a goal: 
\ba{
\theta_{t+1} \leftarrow \theta_t-&\gamma_t (\nabla \mathrm{quality}  
\notag\\
&
~~~~+\lambda \nabla \mathrm{computation}  \left(\theta_t\right)), 
\label{eq:our_pareto_2}
}
where $\nabla \mathrm{computation}$ and $\nabla \mathrm{quality}$ are estimated by score function, and the $\lambda$ can be computed as 
$
\lambda=\max \left(\frac{\phi\left(\theta_t\right)-\nabla \mathrm{quality}\left(\theta_t\right)^{\top} \nabla \mathrm{computation} \left(\theta_t\right)}{\left\|\nabla \mathrm{computation} \left(\theta_t\right)\right\|^2}, \quad 0\right),
$ where $\phi (\theta_t)$ equals to $q(\theta_t)-c$ and the $c$ represents the minimal loss.

\begin{algorithm}[t]
\caption{Switchable Decision ({\ours})} 
\label{alg:switch_decision}
\begin{algorithmic}[1]
\STATE \textbf{Input:} Text $o$. Auto-regressive generation model $\mathcal M$ parameter $w$ with learning rate $\alpha_t$, policy network parameter $\theta$ with learning rate $\gamma_t$, number of iterations $T$.
\FOR{ $t =0$  to $T$}
\STATE $w \leftarrow w - \alpha_t \nabla(w)$, 
\STATE $\theta$ is updated via Eqn \eqref{eq:our_pareto_2},
 \ENDFOR
\end{algorithmic}
\end{algorithm}

\paragraph{The Proposed Algorithm.} Our switchable decision ({\ours}) with efficient candidate space and constrained optimization is 
shown in Algorithm \ref{alg:switch_decision}.
We iteratively update the auto-regressive model and the policy network in a single-loop manner.
The policy network parameter $\theta$ is updated by Eqn~\eqref{eq:our_pareto} in a direction to balance the optimization of quality and constraint satisfaction on computation.

\section{Experimental Settings}\label{sec:experiemental_settings}
Table \ref{tab:dataset_setting} shows the  experimental data configuration. 

\subsection{Task and Evaluation Metrics}
\paragraph{Summarization.}
We use CNN/DailyMail \cite{hermann2015teaching} and XSum \cite{narayan2018don} to evaluate our method. CNN/DailyMail consists of 287,226 documents for training, 13,368 documents
for validation, and 11,490 documents for testing. 
XSum has 226,711 news articles accompanied with a one-sentence summary, answering the question “What is this article about?”. 
Following the splits of \citet{narayan2018don}, it contains 204,045 train, 11,332 dev, and 11,334 test. 
Following
prior work \cite{lewis2019bart}, we use ROUGE \cite{lin2003automatic}
as our primary metric.
We report the unigram ROUGE1 (R-1) and bigram ROUGE-2 (R-2) overlap to assess the informativeness, and the longest common subsequence ROUGE-L (R-L) score to assess the fluency.

\paragraph{Question Answering.}
The Stanford Question Answering Datasets (SQuAD) v1.1 and v2.0 \cite{rajpurkar2016squad,rajpurkar2018know, fan2020bayesian} are popular machine reading comprehension benchmarks. 
For the SQuAD v2.0 dataset, it contains examples where the answer to the question cannot be derived from the provided context.  
Similar to previous settings \cite{devlin2018bert, lewis2019bart}, we use concatenated question and context as input to the encoder of
BART, and additionally pass them to the decoder. 
We report Exact Match (EM) and  F1 score for evaluation \cite{lewis2019bart}.

\paragraph{Classification.}
The General Language Understanding Evaluation (GLUE) benchmark is a collection of natural language understanding (NLU) tasks. 
As shown in Table \ref{tab:dataset_setting}, we include Multi-Genre NLI (MNLI; \cite{williams2017broad, zhang2021learning}), Recognizing Textual Entailment (RTE; \cite{dagan2005pascal}), and Stanford Sentiment Treebank (SST; \cite{socher2013recursive}). 
The diversity of the tasks
makes GLUE very suitable for evaluating the generalization and robustness of our proposed method \cite{liu2020microsoft}. Accuracy is adopted as our evaluation metric.

\begin{table}[h]
\centering
\resizebox{1.0\columnwidth}{!}{
 \begin{tabular}{l|c|c|c|c}
 \toprule
 \bf{Task} &\bf{Dataset} &   \bf{Train} & \bf{Val}  & \bf{Test} \\ \midrule
   \multirow{2}{*}{Summarization} & CNN/DailyMail & 287.2K & 13.4K & 11.5k \\
 & XSum & 204K & 11.3K & 11.3K \\
\hline 
 \multirow{2}{*}{Question Answering} & SQuAD 1.1 & 87.6K & 10.5K & 9.5k \\
 & SQuAD 2.0 & 130.3K & 11.9K & 8.9K \\
  \hline 
 \multirow{3}{*}{Classification} & RTE & 2.5K & 276 & 3k \\
 & MNLI  & 393K & 20K & 20K \\
 & SST & 67K & 872 & 1.8K \\
 \bottomrule
  \end{tabular}}
 \caption{Dataset Configuration. The top block is for summarization, the middle block is for question answering, and the bottom block is the classification tasks.}
    \label{tab:dataset_setting}
\end{table}

\subsection{Implementation Details}\label{sec:implementation_details}

Following \citet{lewis2019bart}, we take the pre-trained BART model as the backbone and utilize the provided checkpoint for finetuning on the downstream datasets. BART is a pre-trained sequence-to-sequence model based on the masked source input and auto-regressive target output, which contains 12 layers of transformer encoder and 12 layers of transformer decoder. Its embedding size is 1,024 and feed-forward size is 4,096. 
We follow the hyper-parameters used in \citet{lewis2019bart}.
Specifically, in summarization, 
we set the training steps as 50k and the number of warm-up steps as 500.
The max number of tokens and the update frequency are set to be 2,048 and 4, respectively. The learning rate is set to $3 \times 10^{-5}$. 
For the question answering (SQuAD 1.1/2.0). We set the total number of updates and warm-up updates as 5,430 and 326, respectively. The max number of sentences is 3 per device with an update frequency of 2. The learning rate is $1.5 \times 10^{-5}$.
We refer the readers to  Appendix \ref{sec:app_exp} for classification hyper-parameter configurations, and
more details about the settings.

\section{Experiments}\label{sec:experiemental_results}
We evaluate the performance of our switchable dynamic network.
In each table, we bold the best result within each column block and the results of our method are obtained with three trials
to determine the variance. See Appendix~\ref{sec:app_exp} for full results with error bars.

\subsection{Summarization}\label{sec:summarization_section}

Table~\ref{tab:summarization_results} reports our results on two summarization datasets. \ding{192} The top block displays the performance of baselines on CNN/DailyMail and XSum datasets, and the bottom block shows the results of incorporating the switchable dynamic networks. We report the results upon the BART large setting in \citet{lewis2019bart}. 
\ding{193}
Summaries in the CNN/DailyMail tend to resemble
source sentences and summaries in XSUM are highly abstractive. 
Baseline models such as BART \cite{lewis2019bart},  UniLM \cite{dong2019unified}, and BERTSUM \cite{liu2019text} do well enough, and even the baseline of the first-three source sentences is highly competitive for CNN/DailyMail. 
Our method can reduce the computation cost while having little or no drop on ROUGE. For example, we even have a 0.2 increase on R1 for CNN/DailyMail and a 0.1 increase on R1 for XSum, while reducing 39\% and  18\% computation costs, respectively.
For the quality of the sentence generations, our method has almost outperformed all the baselines. 
Especially, for the CNN/DailyMail, we achieve better ROUGE with less than two-thirds FLOPs cost, compared to the original BART-large model (\emph{e.g.}, R1: $44.16 \rightarrow 44.31$, RL: $40.90 \rightarrow 41.01$ on CNN/DailyMail).
\ding{194}
These results further confirm that {\ours} can work as an effective module to be incorporated into the auto-regressive generation models. 
{\ours} on improving the inference can also be seen as a complementary module to works focusing on improving pre-training components \citep{hou2022token, ge2022lossless}.

\begin{table}[th]
\centering
\scalebox{0.6}{
 \begin{tabular}{l|cccc|cccc}
 \toprule
\multirow{2}{*}{Model}  & \multicolumn{4}{c|}{CNN/DailyMail} & \multicolumn{4}{c}{XSum}\\ 
  \cline{2-9}
 & R1 $\uparrow$ &  R2 $\uparrow$  & RL $\uparrow$ & FLOPs (\%) $\downarrow$  &  R1 $\uparrow$&  R2 $\uparrow$ & RL $\uparrow$ & FLOPs (\%)  $\downarrow$ \\
   \hline
Lead-3 & 40.42 & 17.62 & 36.67 & - & 16.30 & 1.60 & 11.95 &  -\\
 UniLM& 43.33 & 20.21 & 40.51 &- &-  & - & -  &- \\
BERTSUM  & 42.13 & 19.60 & 39.18 & - &38.81 & 16.50 & 31.27 & - \\
 BART & 44.16 & {\bf 21.28} & 40.90 & 100 & 45.14 & {\bf  22.27} & 37.25 & 100 \\ 
  \hline
 {\bf Ours large} & {\bf 44.31} & {21.18} & {\bf 41.01} & {\bf 61.1}  & {\bf 45.20} & {22.16} & {\bf 37.30} & {\bf 81.9}     \\
 \bottomrule
  \end{tabular}}
 \caption{Comparison to models on CNN/DailyMail and XSum. ROUGE are reported for each model. `BART' represents the BART large model.
 }
    \label{tab:summarization_results}
\end{table}

\paragraph{Comparison with inference reduction methods.}

We adopt several methods from  the conventional early-exiting method (CALM; \cite{schuster2022confident}),  Fast and Robust EarlyExiting (FREE) \cite{bae2023fast}, Pegasus \cite{shleifer2020pre} (pruning and distillation) and DQ-Bart \cite{li2022dq} (quantization and distillation) and compare them with Ours (SD). \ding{182} In \citet{shleifer2020pre}, it utilizes the shrink and finetune methods: BART-student, Pegasus, and BART on CNN/DailyMail. 
\ding{183} In \citet{li2022dq}, it uses quantization and distillation. It reports the BART (8-8-8 6-1). The number at here represents the number of bits for
weights, word embedding, activations, the number of encoder layers, and the number of decoder layers. The results shown in Table \ref{tab:other_inference_reductions} demonstrate our switchable decision achieves a good trade-off between quality and computation. These results verify that our method contributes to efficiency and accuracy, helping the generation model to choose the inference path and candidates dynamically.
\ding{184} Further, combining quantization or distillation method, our effective method of improving the language generation model can also be seen as a complementary and plug-in module. We leave this as a future work. 

\begin{table}[h]
\centering
 \resizebox{0.9\columnwidth}{!}{\begin{tabular}{l|c|c}
 \toprule
Data  & \multicolumn{1}{c|}{ROUGE-L} &\multicolumn{1}{c}{FLOPs (\%)}\\ 
  \cline{2-3}
   \hline
BART-student  & 41.01 & 93.1  \\ 
Pegasus  & 40.34 & 93.1 \\
DQ-Bart  & 40.05 & 18.2  \\
CALM & 40.54 & 80.5 \\
FREE & 40.69 & 76.8 \\
Our Switchable Decision & 41.01 & 61.1 \\ 
 \bottomrule
  \end{tabular}}
 \caption{Comparison {\ours} with different inference cost reduction methods on CNN/DailyMail.  
 }
    \label{tab:other_inference_reductions}
\end{table}

\vspace{-.4em}
\subsection{Classification}\label{sec:classification_section}
We further show the experimental results on the GLUE in Table \ref{tab:results_glue}. The Multi-Genre NLI (MNLI; \cite{williams2017broad}), Recognizing Textual Entailment (RTE; \cite{dagan2005pascal}), and Stanford Sentiment Treebank (SST; \cite{socher2013recursive}) are included. We adopt several baselines from the existing literature.  \ding{172} For BERT, following \citet{bert}, it introduces masked language modeling, which allows pre-training to learn interactions between left and right context words.
\ding{173} UniLM \cite{dong2019unified}, the baseline, fine-tunes BERT with an
ensemble of masks, some of which allow only leftward
context. \ding{174} RoBERTa, following \citet{Liu2019RoBERTaAR}, is pretrained with dynamically changing the mask. \ding{175} For BART \cite{lewis2019bart}, it is a bi-directional encoder-decoder structure.

Table \ref{tab:results_glue} first displays that {\ours} yields a better trade-off between accuracy and computational efficiency.
Ours shows comparable performance over BART and a clear-margin gain over other baselines, while sufficiently lower FLOPs.
For example, {\ours} achieves 87.2\% accuracy \emph{v.s.} BART's 87.0\% accuracy with only 83.6\% FLOPs. 
For the various GLUE benchmarks, our dynamic network demonstrates the strong capability of making skipping decisions for auto-regressive generation models. It further verifies that our method can work for different datasets and can generalize to different input types and fields.

\begin{table}[h]
\centering
\resizebox{1.0\columnwidth}{!}{
 \begin{tabular}{l|cc|cc|cc}
 \toprule
  \multirow{2}{*}{Model}  & \multicolumn{2}{c|}{MNLI} & \multicolumn{2}{c|}{RTE} & \multicolumn{2}{c}{SST}\\ 
  \cline{2-7}
  & m/mm  $\uparrow$ &  FLOPs (\%) $\downarrow$  & Acc $\uparrow$ & FLOPs (\%) $\downarrow$  &  Acc  $\uparrow$ &  FLOPs (\%) $\downarrow$ \\
   \hline
 BERT & 86.6/- & - & 70.4 & - & 93.2 & -\\
 UniLM & 87.0/85.9 & - & 70.9 & - & 94.5 & - \\
  RoBERTa &  90.2/90.2 & - &  86.6 & - & 96.4 & -  \\
  BART & {\bf 89.9/90.1} & 100 & 87.0 & 100 & {\bf 96.6} & 100   \\
\cline{1-7}
 {\bf Ours} & { 89.7/90.0} & {\bf  82.4} & {\bf 87.2} & {\bf 83.6} & {\bf 96.6} & {\bf 80.7}  \\
 \bottomrule
  \end{tabular}}

 \caption{Performance on GLUE. We report the accuracy. All language models here are large size. `m/mm' and `Acc' denotes accuracy on matched/mismatched version MNLI and accuracy, respectively.
 }
    \label{tab:results_glue}
\end{table}

\subsection{Question Answering}\label{sec:question_answering_section}
For both SQuAD v1.1 and v2.0, following \citet{lewis2019bart}, we feed the complete documents into the encoder and decoder, and use the top hidden state of the decoder as a representation for each
word. This representation is used to classify the token.
Table~\ref{tab:results_question_answering} shows our experiment results. The BART large is used as the
primary baseline, and the recent baselines \cite{bert, dong2019unified, liu2019roberta} are reported. 
We load the official checkpoint from $\mathrm{Fairseq}$ with the official pre-processed SQuAD data.
On question answering, by dynamically skipping the candidates from attention layers, feed-forward layers, and input tokens, our model achieves a similar EM and F1 score as BART. 
Different from the above tasks, here the input is concatenated question and context and additionally passed to the decoder.
Although the input is organized in different formats, it is interesting to see the consistent computation cost improvement of our proposed switchable decision in question answering. It further demonstrates that {\ours} can be utilized in general NLP tasks. 
\begin{table}[h]
\centering
\resizebox{1.0\columnwidth}{!}{
 \begin{tabular}{l|cc|cc}
 \toprule
  \multirow{2}{*}{Model}  & \multicolumn{2}{c|}{SQuAD 1.1} & \multicolumn{2}{c}{SQuAD 2.0} \\
  \cline{2-5}
  & EM/F1 $\uparrow$ &  FLOPs (\%) $\downarrow$  & EM/F1 $\uparrow$ & FLOPs (\%) $\downarrow$   \\
   \hline
 BERT & 84.1/90.9 & - & 79.0/81.8 & - \\
 UniLM & -/- & - & 80.5/83.4 & - \\
  RoBERTa &  {\bf 88.9}/{\bf 94.6} & - &  {\bf 86.5}/{\bf 89.4} & - \\
  BART & 88.8/{\bf 94.6} & 100 & 86.1/89.2 & 100    \\
\cline{1-5}
 {\bf Ours} & {88.7/94.5} & {\bf 80.5} & { 86.0/89.3} & {\bf 83.3}   \\
 \bottomrule
  \end{tabular}}
 \caption{Results across different strategies on SQuAD v1.1 and v2.0. Answers are text spans extracted from a given document context.
\footnotemark 
 }
    \label{tab:results_question_answering}
\end{table}

\section{Analysis}\label{sec:analysis}
\paragraph{Can we use the proposed dynamic network with the different auto-regressive generation models?} 

As discussed in Section \ref{sec:method_section}, our proposed method targets the auto-regressive generation model. Thus, can our method be adapted to other auto-regressive generation models? We select the GPT-2 \cite{radford2019language} base and T5 \cite{Raffel2020ExploringTL} base to study the performance after adapting our proposed switchable decisions. The results are presented in Table \ref{tab:analysis_t5_and_bart}.
It indicates our method is insensitive to different generation models.
This confirms our discussion in Section \ref{sec:method_section} that  {\ours} can serve as an efficient alternative dynamic network for versatile generation models. We also analyze the impact of making decisions based on different hidden representations. More details about LLaMA \cite{touvron2023llama} models are included in Appendix~\ref{sec:app_exp}.

\begin{table}[h]
\centering
 \resizebox{0.7\columnwidth}{!}{\begin{tabular}{l|c|c}
 \toprule
Data  & \multicolumn{1}{c|}{ROUGE} &\multicolumn{1}{c}{ FLOPs (\%)}\\ 
  \cline{2-3}
   \hline
BART & 44.16/21.28/40.90 & 100 \\  
+ Ours & 44.31/21.18/41.01 & 61.1 \\
\hline
GPT-2 & 37.55/15.53/25.81 & 100 \\ 
+ Ours & 37.76/15.68/25.93 & 74.5 \\
\hline
T5 & 42.05/20.34/39.40 & 100 \\ 
+ Ours & 41.98/20.38/39.61 & 74.5 \\
 \bottomrule
  \end{tabular}}
 \caption{The proposed method for different generation models on CNN/DailyMail. 
 }
    \label{tab:analysis_t5_and_bart}
\end{table}


\paragraph{What are the differences between encoder-only, decoder-only, and token-only architecture search space?} \label{analysis:encoder_decoder}
We test if our results are sensitive to the choice of architectures: encoder-only, decoder-only, and encoder-decoder. We create the following scenarios: \ding{172} For encoder-only, we incorporate the attention and feed-forward as the skipping candidates. \ding{173} For decoder-only, similarly, the attention and feed-forward are included. \ding{174} For token-only, the token candidate is utilized. Then we compare these three designs with {\ours} and BART large to see the impact of incorporating our designed decision space into these different model architectures. As shown in Table \ref{tab:analysis_encoder_decoder}, we observe distinct FLOPs (reducing 10\%) saving by only adding our skipping attention and feed-forward strategies for encoder-only and decoder-only. By only including the token skipping for the encoder-decoder structure, we observe the larger FLOPs (reducing 29\%) saving while delivering the comparable ROUGE to BART. We refer the readers to Appendix \ref{sec:contributions_of_search_space_candidates} for the detailed skipping percentage of each candidate. 
These results confirm our analysis and motivation for the switchable decision that using a combination of all these architectural search spaces comes to the best efficiency and accuracy trade-off. 

\begin{table}[h]
\centering
 \resizebox{1.0\columnwidth}{!}{\begin{tabular}{l|c|c|c|c|c}
 \toprule
 Architecture & ATT & FFN & Token & FLOPs (\%) & ROUGE  \\ \midrule
 BART &   &  & & 100 & 44.16/21.28/40.90 \\
  \hline
 Encoder-Only &\cmark & \cmark& & 91.9 & 44.21/{\bf 21.32}/40.95  \\
 Decoder-Only & \cmark & \cmark & & 90.3 & 44.13/21.08/40.86 \\
   Token-Only & & & \cmark &  71.5 & 44.09/21.26/40.92  \\
   \hline
   Ours &  \cmark & \cmark &\cmark &  {\bf 61.1} & {\bf 44.31}/{ 21.18}/{\bf 41.01} \\
 \bottomrule
  \end{tabular}}
 \caption{Results of skipping strategies on different architecture spaces for CNN/DailyMail. BART \cite{izacard2021distilling} large model is presented. 
} 
    \label{tab:analysis_encoder_decoder}
\end{table}

\paragraph{Ablation studies on the components in {\ours}.} 
We conduct the ablation study to examine the role of constrained optimization. 
For ablation, instead of automatically searching the trade-off between the quality and computation,
we manually set the $\lambda$ in Eqn \eqref{eq:our_pareto_2} as $0.2$, $0.5$, $0.8$. 
We also include the random selection strategy. The random selection strategy is not learning switchable decisions and would not dynamically assign computation for each data instance.
\ding{182} Table \ref{tab:component_analysis} shows that 
the constrained optimization of our method brings clear benefits. \ding{183} We find that without CO, `$\minus$ CO' with different manually tuned $\lambda$ value shows an unstable trade-off between the ROUGE and FLOPs across all $\lambda$ values, indicating that manually tuned $\lambda$ value can not bring both optimized quality and computation together.
\ding{184}Empirically, we randomly select a policy from our decision space candidates and use the same other parameters. These result in a degradation in performance and lower FLOPs reduction.
It demonstrates the necessity and effectiveness of the constrained optimization for the switchable candidate set in {\ours} structure.

\begin{table}[h]
\centering
 \resizebox{0.7\columnwidth}{!}{\begin{tabular}{l|c|c}
 \toprule
Data  & \multicolumn{1}{c|}{ROUGE} &\multicolumn{1}{c}{ FLOPs (\%)}\\ 
  \cline{2-3}
   \hline
BART & 44.16/21.28/40.90 & 100 \\  
Random & 41.77/19.02/38.72 & 75.3 \\
Ours & 44.31/21.18/41.01 & 61.1 \\ 
- CO, $\lambda$ = 0.2 & 44.12/21.30/40.88 & 77.8 \\
- CO, $\lambda$ = 1.0 & 42.89/21.02/40.57 & 68.5 \\
- CO, $\lambda$ = 1.5 & 41.35/19.87/38.39 & 49.4 \\
 \bottomrule
  \end{tabular}}
 \caption{Comparison of different $\lambda$ values for the manually tuned trade-off between computation and quality vs. Ours. `CO' denotes constrained optimization.
 }
    \label{tab:component_analysis}
\end{table}

\paragraph{Efficiency and time.} 
We provide the parameter sizes, average GPU memory per device, per step training time, and inference time comparisons between the baseline and {\ours} during the finetuning. Experiments in this part
are performed on eight Tesla V100 GPUs. 
\ding{182}
Table \ref{tab:runningtime} shows that {\ours} keeps the parameter size at the same level as the BART large during finetuning. The GPU memory per device and training time of {\ours} are slightly higher (2.7\% for memory and 1.6\% for running time) than BART. 
{\ours} gives the best inference FLOPs, outperforming BART while keeping the comparable ROUGE score and running time. \ding{183} 
For the inference time, we evaluate our method and BART large on CNN/DailyMail following the same setting and device with batch size 1. For each iteration, 5.1 seconds (Ours) vs. 10.3 seconds (BART). Our dynamic network demonstrates the strong capability of making skipping decisions.
\ding{184}
With the constrained optimization and the reinforcement learning agent, our switchable decision is still computationally productive as the design of our optimization and agent (\emph{e.g.,} applying one-layer MLP for policy network) has almost negligible finetuning computational cost.

\begin{table}[h]
\centering
\resizebox{1.0\columnwidth}{!}{\begin{tabular}{l|c|c|c|c|c}
 \toprule
 Model & ROUGE $\uparrow$ & Params $\downarrow$ & GPU memory $\downarrow$ & s/step $\downarrow$ & IT  $\downarrow$\\ 
 \hline
BART & 44.16/21.28/40.90 & 406M & 16.8G   & 1.20  & 10.3\\ 
Ours  & 44.31/21.18/41.01 & 423M  & 17.6G & 1.48 & 5.1\\
 \bottomrule
  \end{tabular}}
 \caption{Results of parameter size, GPU memory per device, and step time for BART and ours finetuning on CNN/DailyMail. `s/step' represents training step time (second/per step).`IT' represents inference time (second) for each iterations.}
\label{tab:runningtime}
\end{table}

\subsection{Contributions of Search Space Candidates.}\label{sec:contributions_of_search_space_candidates}
To further identify the contributions of our search space candidates for efficiency improvements and inference acceleration, we present the details skipping percentage of each candidate for CNN/DailyMail, SQuAD 1.1, and SST in Table \ref{tab:contribution_candidate_flops}. For CNN/DailyMail, we observe around $8\%$ attention skipping of total attention, $11\%$ feed-forward skipping of total feed-forward, and $29\%$ token skipping of total tokens. 
The similar skipping percentage holds for question answering. 
However, we have seen an obvious contrast in the token skipping percentage in classification tasks.
The key observation is that the skipping percentages for tokens are high for both CNN/DailyMail and SQuAD 1.1. 
In addition, our method generally takes around 5K iterations for the reinforcement learning algorithm to converge on CNN/DailyMail. 
This confirms our conjecture in Section \ref{sec:summarization_section}. 
For summarization and question answering tasks, 
the first few parts of inputs are more representative. 
Thus, it perfectly serves as the candidate for our switchable network to make the skipping decisions. 

\begin{table}[h]
\centering
\resizebox{0.7\columnwidth}{!}{
 \begin{tabular}{l|c|c|c}
 \toprule
\bf{Dataset} &   \bf{ATT} & \bf{FFN}  & \bf{Token} \\ \midrule
   CNN/DailyMail & 8.50\%& 11.13\% & 28.75\% \\
\hline 
SST &  13.30\%& 13.54\% & 7.18\%  \\
  \hline 
 SQuAD 1.1 & 10.21\%& 11.93\% & 9.02\% \\
 \bottomrule
  \end{tabular}}
 \caption{Skipping percentage of each candidate. For example, 8.50\% indicates that there are 8.50\% of total attention skipped.
}
    \label{tab:contribution_candidate_flops}
\end{table}

\paragraph{The impact of making decisions based on different hidden representations.}
In Section \ref{sec:construct_decision_space}, we consider three skipping candidates' hidden representations (attention, feed-forward, and query) after the first layer as the input for our reinforcement learning agent to make switchable decisions.
Here, we demonstrate that using hidden representations from different layers comes to the same results, and therefore we pick the easiest one.
We set up a baseline here, in which whether to skip the following layer is dependent on the nearby previous layer outputs.
We experiment on Ours (based on the output from the first layer) and Ours Layer Wise (layer-wise decisions based on the output from the nearby previous layers). 
The difference between these two cases is small in Table \ref{tab:layer_wise_decisions}. 
The layer-wise design requires more computation as it needs to make decisions at each layer. 
Therefore, it further demonstrates that the design of ours is capable of making skipping decisions and imposing less computational cost. 

\begin{table}[h]
\centering
 \resizebox{0.83\columnwidth}{!}{\begin{tabular}{l|c|c}
 \toprule
Data  & \multicolumn{1}{c|}{ROUGE} &\multicolumn{1}{c}{FLOPs (\%)}\\ 
  \cline{2-3}
   \hline
Ours & 44.31/21.18/41.01 & 61.1  \\  
Ours Layer Wise & 44.38/21.22/40.97 & 61.8 \\ 
 \bottomrule
  \end{tabular}}
 \caption{Comparison of different layer-wise decision of {\ours} on CNN/DailyMail. `Ours' represents the decision based on the hidden after the first layer. `Ours Layer Wise' represents the decision based on the hidden representation from the nearby previous layer. 
 }
    \label{tab:layer_wise_decisions}
\end{table}

\section{Conclusion}\label{sec:conclusion}
Our work demonstrates the benefits of introducing a switchable decision of the dynamic network. The proposed method can dramatically increase the inference efficiency and still enable the model performance. 
Noticeable FLOPs saving and consistent performance are observed across summarization, question answering, and classification benchmarks. We further conduct a detailed study with the proposed switchable strategy in different settings, \emph{e.g.}, comparing with different architecture search spaces, providing more evidence for making decisions based on hidden representations, and verifying the impact of components.
To summarize, the proposed {\ours} is effective and general, with the potential to be incorporated into existing generation models for various NLP tasks.


\clearpage
\bibliographystyle{icml2024}
\bibliography{reference}

\clearpage

\appendix
\section{Experimental details}\label{sec:app_exp}
\subsection{Full Results With Error Bar }\label{sec:appendix_fullresults}
We report the full results of our method with the error bar for summarization and question answering in Table \ref{tab:appendix_summarization_results} and  \ref{tab:appendix_results_question_answering}, respectively. The full result of classification is demonstrated in Table \ref{tab:appendix_results_glue}.

\begin{table}[th]
\centering
\scalebox{0.4}{
 \begin{tabular}{l|cccc|cccc}
 \toprule
\multirow{2}{*}{Model}  & \multicolumn{4}{c|}{CNN/DailyMail} & \multicolumn{4}{c}{XSum}\\ 
  \cline{2-9}
 & R1 $\uparrow$ &  R2 $\uparrow$  & RL $\uparrow$ & FLOPs (\%) $\downarrow$  &  R1 $\uparrow$&  R2 $\uparrow$ & RL $\uparrow$ & FLOPs (\%)  $\downarrow$ \\
   \hline
Lead-3 & 40.42 & 17.62 & 36.67 & - & 16.30 & 1.60 & 11.95 &  -\\
 UniLM& 43.33 & 20.21 & 40.51 &- &-  & - & -  &- \\
BERTSUM  & 42.13 & 19.60 & 39.18 & - &38.81 & 16.50 & 31.27 & - \\
 BART & 44.16 & {21.28} & 40.90 & 100 & 45.14 & {22.27} & 37.25 & 100 \\ 
  \hline
 {Ours large} & {44.31$\pm$0.1} & {21.18$\pm$0.2} & { 41.01$\pm$0.2} & { 61.1}  & { 45.20$\pm$0.1} & {22.16$\pm$0.2} & {37.30$\pm$0.2} & { 81.9}     \\
 \bottomrule
  \end{tabular}}
 \caption{Full results on CNN/DailyMail and XSum. ROUGE is reported for each model. `BART' represents the BART large model. 
 }
    \label{tab:appendix_summarization_results}
\end{table}

\begin{table}[h]
\centering
\resizebox{1.0\columnwidth}{!}{
 \begin{tabular}{l|cc|cc|cc}
 \toprule
  \multirow{2}{*}{Model}  & \multicolumn{2}{c|}{MNLI} & \multicolumn{2}{c|}{RTE} & \multicolumn{2}{c}{SST}\\ 
  \cline{2-7}
  & m/mm  $\uparrow$ &  FLOPs (\%) $\downarrow$  & Acc $\uparrow$ & FLOPs (\%) $\downarrow$  &  Acc  $\uparrow$ &  FLOPs (\%) $\downarrow$ \\
   \hline
 BERT & 86.6/- & - & 70.4 & - & 93.2 & -\\
 UniLM & 87.0/85.9 & - & 70.9 & - & 94.5 & - \\
  RoBERTa &  90.2/90.2 & - &  86.6 & - & 96.4 & -  \\
  BART & { 89.9/90.1} & 100 & 87.0 & 100 & { 96.6} & 100   \\
\cline{1-7}
 { Ours} & { 89.7$\pm$0.2/90.0$\pm$0.3} & {82.4} & {87.2$\pm$0.1} & {83.6} & { 96.6$\pm$0.2} & { 80.7}  \\
 \bottomrule
  \end{tabular}}
 \caption{Full performance on GLUE. We report the accuracy of each dataset. All language models here are large size. `m/mm' and `Acc' denotes accuracy on matched/mismatched version MNLI and accuracy, respectively.
 }
    \label{tab:appendix_results_glue}
\end{table}

\begin{table}[h]
\centering
\resizebox{1.0\columnwidth}{!}{
 \begin{tabular}{l|cc|cc}
 \toprule
  \multirow{2}{*}{Model}  & \multicolumn{2}{c|}{SQuAD 1.1} & \multicolumn{2}{c}{SQuAD 2.0} \\
  \cline{2-5}
  & EM/F1 $\uparrow$ &  FLOPs (\%) $\downarrow$  & EM/F1 $\uparrow$ & FLOPs (\%) $\downarrow$   \\
   \hline
 BERT & 84.1/90.9 & - & 79.0/81.8 & - \\
 UniLM & -/- & - & 80.5/83.4 & - \\
  RoBERTa &  { 88.9}/{ 94.6} & - &  { 86.5}/{ 89.4} & - \\
  BART & 88.8/{ 94.6} & 100 & 86.1/89.2 & 100    \\
\cline{1-5}
 { Ours} & { 88.7$\pm$0.3/94.5$\pm$0.4} & { 80.5} & { 86.0$\pm$0.3/89.3$\pm$0.3} & { 83.3}   \\
 \bottomrule
  \end{tabular}}
 \caption{Full results across different strategies on SQuAD v1.1 and v2.0. Answers
are text spans extracted from a given document context.
 }
    \label{tab:appendix_results_question_answering}
\end{table}



\subsection{Experimental Datasets}
\paragraph{Summarization.}
CNN/DailyMail contains news articles and associated highlights as summaries. 
Following the standard splits from \citet{hermann2015teaching} for training, validation, and testing, we have 90,266/1,220/1,093 CNN
documents and 196,961/12,148/10,397 DailyMail
documents, respectively. The sentence is split by using the Stanford CoreNLP toolkit \cite{manning2014stanford}. For XSum \cite{narayan2018don}, summaries are professionally written by the authors of the documents. We also use the pre-processing and data splits from \cite{narayan2018don, yang2024preference}.

\paragraph{Question Answering.}
Stanford Question Answering Dataset (SQuAD) \cite{rajpurkar2016squad, rajpurkar2018know, zhang2021knowing, zhang2022passage} is an extractive question answering task, consisting of questions posed by crowdworkers on a set of Wikipedia articles. The answers, given the questions, are text span from the given reading passage. The SQuAD 1.1 contains around 100,000 question-answer pairs on about 500 articles.
The SQuAD v2.0 dataset includes unanswerable questions about the same paragraphs.

\paragraph{Classification.}
GLUE \cite{wang2018glue, zhang2022allsh} comprises a collection of text classification tasks meant to test general language understanding abilities. We adopt the three datasets for our experiments: natural language inference (MNLI \cite{Williams2017ABC} and RTE \cite{dagan2005pascal}) and sentiment analysis (SST-2 \cite{socher2013recursive}).  

\subsection{Experimental Settings}
For summarization, we follow the setting in \cite{lewis2019bart} and initialize our models with the pretrained BART large checkpoint. The checkpoint is from the Fairseq library \footnote{\url{https://github.com/facebookresearch/fairseq/tree/main/examples/bart}}. 
T5 \cite{Raffel2020ExploringTL} is also used in Section \ref{sec:analysis}. We adopt the T5 base from the HuggingFace Transformer library\footnote{\url{https://github.com/huggingface/transformers}}. Following \citet{lewis2019bart}
, the Adam optimizer \cite{kingma2014adam, liu2021fusedream, zhang2024language} is utilized for optimizing the model parameter with the learning rate $3 \times 10^{-5}$. The training step is 50k and the warmup step is 500. Both dropout and attention dropout are set as 0.1.
For classification, the detailed training settings are presented in Table \ref{tab:glue_detailed_setting}. 
\begin{table}[h]
\centering
 \resizebox{0.8\columnwidth}{!}{\begin{tabular}{l|c|c|c}
 \toprule
Model  & \multicolumn{1}{c|}{MNLI} &\multicolumn{1}{c|}{RTE} &\multicolumn{1}{c}{SST-2}\\ 
  \cline{2-3}
   \hline
NC & 3 & 2 & 2  \\  
LR &  $5 \times 10^{-6}$ &  $1 \times 10^{-5}$ &  $5 \times 10^{-6}$\\ 
BSZ & 128 & 32 & 128 \\ 
TS & 30,968 & 1,018 & 5,233\\ 
WS & 1,858 & 61 & 314 \\ 
 \bottomrule
  \end{tabular}}
 \caption{Experiment setting for MNLI, RTE, and SST-2 (LR: learning rate, BSZ: batch size, NC: number of classes, TS: total number of training steps, WS: warm-up steps).
 }
    \label{tab:glue_detailed_setting}
\end{table}

\begin{table}[h]
\centering
 \resizebox{0.6\columnwidth}{!}{\begin{tabular}{l|c|c}
 \toprule
Data  & \multicolumn{1}{c|}{ROUGE} &\multicolumn{1}{c}{ FLOPs (\%)}\\ 
  \cline{2-3}
   \hline
BART & 44.16/21.28/40.90 & 100 \\  
+ Ours & 44.31/21.18/41.01 & 61.1 \\
\hline
GPT-2 & 37.55/15.53/25.81 & 100 \\ 
+ Ours & 37.76/15.68/25.93 & 74.5 \\
\hline
T5 & 42.05/20.34/39.40 & 100 \\ 
+ Ours & 41.98/20.38/39.61 & 74.5 \\
\hline
LLaMA & -/-/46.68 & 100 \\ 
+ Ours & -/-/46.73 & 77.6 \\
 \bottomrule
  \end{tabular}}
 \caption{The proposed method for different generation models on CNN/DailyMail. 
 }
    \label{tab:appendix_t5_and_bart_llama}
    \vspace{-10pt}
\end{table}
\subsection{More comparisons} 

As discussed in Section \ref{sec:analysis},  We select the GPT-2 \cite{radford2019language} base and T5 \cite{Raffel2020ExploringTL} to study the performance after adapting our proposed switchable decisions. We also included LLaMA \cite{touvron2023llama} as an additional comparison in Table \ref{tab:appendix_t5_and_bart_llama}.

\end{document}